\documentclass{article} 
\usepackage{colm2024_conference}

\usepackage{microtype}
\usepackage{hyperref}
\usepackage{url}
\usepackage{booktabs}
\usepackage{amsmath}
\usepackage{amssymb}
\usepackage{mathtools}
\usepackage{amsthm}
\usepackage{wrapfig}

\usepackage[pass]{geometry}

\usepackage[capitalize,noabbrev]{cleveref}
\usepackage{graphicx}
\usepackage{subcaption}
\usepackage{booktabs} 
\usepackage{booktabs}
\usepackage{multirow}
\usepackage{siunitx}
\usepackage{adjustbox}
\usepackage{threeparttable}
\usepackage[table]{xcolor}  
\usepackage{arydshln}  
\usepackage[table]{xcolor}  
\definecolor{lightblue}{RGB}{230,230,255} 

\usepackage[T1]{fontenc}
\usepackage{microtype}
\usepackage{setspace}
\usepackage{mathptmx}
\definecolor{darkblue}{rgb}{0, 0.2, 0.8}
\hypersetup{colorlinks=true, citecolor=darkblue, linkcolor=darkblue, urlcolor=darkblue}

\title{{\fontsize{13.5pt}{10pt}\selectfont
 Quantifying and Predicting Disagreement in Graded Human Ratings}}

\author{\hspace{0cm}Leixin Zhang\thanks{\texttt{leixin.zhang@uni-tuebingen.de}} \hspace{0.3cm} Çağrı Çöltekin\thanks{\texttt{cagri.coeltekin@uni-tuebingen.de}} 
\vspace{0.2cm}\\University of Tübingen, Germany
}

\colmfinalcopy 

\begin{document}

\maketitle
\thispagestyle{plain}

\begin{abstract}
It is increasingly recognized that human annotators do not always agree, and such disagreement is inherent in many annotation tasks. However, not all instances in a given task elicit the same degree of opinion divergence. In this paper, we investigate annotation variation patterns in graded human ratings for inappropriate languages, including offensive language, hate speech, and toxic language perception. We examine whether the degree of annotation disagreement can be predicted from textual features. We further propose the \textit{Opposition Index}, a metric that quantifies perspective opposition among annotators on a given item, and investigate the predictability of instances with potentially opposing human opinions.
Our results show a moderate positive correlation between estimated and observed annotation variance. We find that two approaches achieve comparable performance in variance prediction: directly predicting the variance value and estimating it from predicted annotation distributions. Our results on opposition perspective prediction show that items with high opposition index values are more difficult to predict and are often underestimated by models.

\end{abstract}

\section{Introduction}


Recent work has demonstrated that many natural language processing (NLP) datasets and tasks exhibit inherent annotation variation \citep{plank-2022-problem, sorensen2024roadmap}. This variation occurs across multiple linguistic domains and task types. In \textit{{syntax}}, it is observed in annotation tasks such as part-of-speech tagging \citep{plank-etal-2014-learning, zeman2010hard}; in \textit{{semantics}}, such as semantic similarity \citep{wang2023collective} and natural language inference (NLI) \citep{huang2023culturally, jiang-marneffe-2022-investigating, pavlick2019inherent, liu2023we, zhang-de-marneffe-2021-identifying}; in \textit{{pragmatics}}, including irony detection \citep{casola-etal-2024-multipico} and dialogue act annotation \citep{verdonik2023annotating}; and in other \textit{{socially-relevant NLP tasks}}, such as hate speech detection \citep{sang2022origin, macavaney2019hate}, offensive language \citep{kocon2021offensive, davani2024d3code}, and toxic content detection \citep{kumar2021designing}.

\vspace{0.1cm}

It is increasingly acknowledged that human annotators do not always make identical decisions or hold the same opinions on many NLP tasks \citep{plank-2022-problem, uma2021learning,basile2021we}. However, the annotation pattern is not consistent across all items in the same task: some cases are clear-cut and have near-perfect agreement from multiple annotators, while others can be more ambiguous, resulting in variance in annotation patterns.  
Such annotation variation can arise from item ambiguity or vagueness (e.g., insufficient context), language complexity (e.g., use of slang or jargon), or annotators’ personal beliefs, values, expertise, or personality \citep{sap-etal-2020-social,sap-etal-2022-annotators}. 
Reflected in Likert rating distributions, annotations for some items show sharper peaked distributions, indicating strong consensus among annotators, while others may display flatter or multi-modal distributions, reflecting interpretation variability or the presence of opposing opinions among annotators.

\vspace{0.15cm}

Estimating which items are likely to elicit disagreement has important practical and theoretical implications. In \textit{{annotation practice}}, identifying disagreement-prone items allows researchers to optimize annotation workflows by prioritizing difficult or perspective-divergent cases. For instance, socially controversial items such as potentially offensive or politically charged content often require a larger number of annotators to capture the diversity of latent perspectives, whereas less controversial items may require fewer annotations.
\vspace{0.15cm}

From a \textit{{linguistic perspective}}, analyzing uncertainty patterns allows researchers to uncover the latent factors underlying annotation uncertainty, such as detecting cases that lack sufficient context \citep{sandri-etal-2023-dont}, and provide insights into human perception of complex language phenomena, such as irony, sarcasm, or figurative language. Studying perspective disagreement can further reveal culturally dependent interpretations (e.g., Western vs. non-Western perspectives \citep{sap-etal-2022-annotators,huang2023culturally,larimore-etal-2021-reconsidering}, liberal vs. conservative viewpoints \citep{luo-etal-2020-detecting}) or conflicts in judgment.  
In \textit{{other decision-making domains}}, including legal, political, and medical decision-making, annotation variation may reflect conflicting interests and opposing perspectives between different parties (e.g., employers vs. employees, producers vs. consumers) \citep{angouri2012managing,braun2024beg}.
Automatically identifying disagreement-prone items can help flag conflicting cases, prioritize expert review, and improve decision fairness \citep{patel-etal-2018-annotation}.

\vspace{0.15cm}

In this work, we investigate whether the item-level annotation patterns can be inferred solely from item features, and we mainly focus on inappropriate language detection tasks in this work, including hate speech, offensive, and toxic language classification. 
These tasks have been extensively studied and are known to exhibit substantial annotator disagreement \citep{sang2022origin}, and there is a lack of universally accepted standards or definitions. 
For example, definitions of hate speech vary across research objectives \citep{talat2016hateful}, legal frameworks \citep{ec_hate_speech_code_2016}, and platform policies. It is often impractical to specify every possible case in annotation guidelines, particularly in crowdsourced settings where annotators are not formally trained. Thus, hate speech annotation often relies on annotators’ perceptions, linguistic intuition, and individual understanding of what constitutes hate speech. In the era of large language models (LLMs), some challenges become more critical \citep{weidinger2021ethical}. Given the rapidly growing user base of LLM-powered systems, detecting toxic and inappropriate language is essential for mitigating risks, preventing the amplification of harmful content, and ensuring safer user interactions.
\vspace{0.15cm}

To more faithfully capture the nuances of human judgment and annotation distributions,
we use datasets annotated with Likert-scale ratings instead of discrete binary labels in this work. We also tailor the training objective by employing loss functions designed for ordinal data rather than simple discrete classes, including the Earth Mover’s Distance (and its variant) and cumulative cross-entropy. This work focuses on estimating two aspects of human annotation variation:

\vspace{-2pt}

\begin{itemize}
    \item \textbf{Annotation variance:} whether the degree of dispersion in annotator responses can be inferred from item features.  
\vspace{-5pt}
    \item \textbf{Opposing stances:} whether the conflicting stances among annotators can be effectively modeled and predicted.
\end{itemize}
\vspace{-2pt}

\noindent By modeling item-level annotation variance and stance opposition, our work takes a step toward characterizing patterns or structures of human annotation variation. We hope this study will inspire further research on annotation uncertainty and perspective-aware NLP systems.

\section{Related Work}

Existing literature on annotation variation can be broadly divided into two main streams. 
The first stream focuses on analyzing human annotation and human interpretation variation \citep{hong2025agree,jiang-marneffe-2022-investigating}, investigating types or causes of variation across annotators \citep{xu-etal-2023-dissonance}, disentangling noise from genuine disagreement \citep{weber-genzel-etal-2024-varierr}, and analyzing cultural background influence \citep{huang2023culturally}. 
The second stream focuses on modeling human annotation variation with machine learning approaches \citep{uma2021learning,davani-etal-2022-dealing,zhou-etal-2022-distributed}. The methods include soft-label training \citep{uma2021learning,fornaciari2021beyond}, incorporating socio-demographic features for group perspective simulation \citep{gordon2022jury}, multi-task learning \citep{davani-etal-2022-dealing} with each task corresponding to a specific annotator, and using an annotator embedding layer \citep{mokhberian-etal-2024-capturing} to learn annotator-specific labels.

\vspace{0.15cm}

Work that explicitly predicts or infers disagreement from text features remains relatively scarce. In this direction, \citet{zhang-de-marneffe-2021-identifying} aim to tease apart agreed and disagreed items in natural language inference (NLI) and propose an ensemble approach by integrating three specialized models trained to predict three labels: entailment, neutral, or contradiction. In subjective tasks such as hate speech detection, \citet{wan2023everyone} model disagreement by directly predicting whether or not annotation variation exists for an item as a binary classification problem, and a regression problem by predicting annotation variation with the value 
$1 - P_{{majority}}$, the proportion of annotations that do not fall into the majority label. \citet{baumler-etal-2023-examples} and \citet{van-der-meer-etal-2024-annotator} estimate the uncertainty of human annotations and incorporate it into an active learning framework.

\vspace{0.15cm}

Despite these advances, prior work has largely treated opinion divergence as a categorical prediction problem (e.g., three-way NLI labels or binary hate speech decisions) and has not examined the degree of annotation uncertainty and whether the full structure of fine-grained annotation distributions can be recovered from item-level signals \citep{zhang-2025-proposal}. To address this gap, we perform a detailed analysis to model the full distribution of Likert-scale ratings and examine its effectiveness in inferring annotators' opinion divergence with item textual features.

\section{Rating Disagreement Prediction}

To model opinion divergence, including both variance and opposing views, we use datasets with Likert-scale ratings. Unlike discrete labels, these graded ratings capture fine-grained human perception differences, reflect gradations in perceptions of inappropriate language, enabling analysis of multimodal and polarized patterns in distribution.
We aim to test whether the structure of annotation variation across items can be captured from textual features. Specifically, we examine the magnitude of opinion divergence in \cref{variaiton_method} and the presence of opposing stances among annotators in \cref{opposing_method}.

\subsection{Inferring Rating Variance} \label{variaiton_method}

For discrete classes, the entropy of annotator labels is commonly used to quantify uncertainty. For Likert ratings, where the distance between ratings is meaningful, we treat them as equally spaced values. For each item \(i\), the degree of annotation variation is computed as the unbiased variance of the ratings from \(N_i\) annotators:

\begin{equation}
\sigma_i^2 = \frac{1}{N_i - 1} 
\sum_{j=1}^{N_i} (r_{ij} - \bar{r}_i)^2,
\end{equation}

\noindent where \(r_{ij}\) is the rating of annotator \(j\) on item \(i\), and \(\bar{r}_i\) is the mean rating for that item. Perfect agreement corresponds to \(\sigma_i^2 = 0\).

\vspace{10pt}
\noindent To estimate annotation variation, we consider two approaches:

\begin{enumerate}
    \item \textbf{Direct variance prediction}: we train a regression model to predict the unbiased variance \(\sigma_i^2\) directly from item-level features.
    \item \textbf{Distribution-based prediction}: we predict the full Likert rating distribution \(\mathbf{p}_i = [p_{i1}, \dots, p_{iK}]\) for item \(i\), then compute the variance from the predicted distribution:
    \begin{equation}
    \hat{\sigma}_i^2 = \sum_{k=1}^K p_{ik} (k - \hat{\mu}_i)^2, \quad 
    \hat{\mu}_i = \sum_{k=1}^K p_{ik} \cdot k.
    \end{equation}
\end{enumerate}

\noindent To evaluate the performance of prediction, we mainly measure with the following metrics: (1) \textit{Mean Squared Error (MSE)} between predicted and true variance, (2) \textit{Spearman's rank correlation $r$} between predicted variance and true variance across items and (3) $F_1$ score of whether or not opinion divergence or rating difference is in the annotation of an item among annotators.

\subsection{Identifying Opposing Opinions}\label{opposing_method}

\begin{wraptable}{l}{0.5\textwidth}{}
\centering
\footnotesize
\renewcommand{\arraystretch}{2.2} 
\begin{tabular}  {p{0.02\linewidth} p{0.5\linewidth} p{0.28\linewidth}}
\toprule
 & \textbf{Comments} & \textbf{Annotations} \\ \hline

1 & \textquotedblleft Mr \#Trump will be loving today. As it is the one day of the year when \#FakeNews is acceptable. \#aprilfools \textquotedblright
& [0, 0, 0, 2, 2, 2] \\ \hline

2 & \textquotedblleft Nigga at da end of the day we all would be gone, or somewhere else. and speakin about it is not gonna fucking matter! \textquotedblright
& [0, 0, 2, 2, 2] \\ \hline

3 & \textquotedblleft I hate when guys call their girls bitches and hoes. That's your girl. You respect her. \textquotedblright
& [0, 0, 0, 2, 2] \\ \bottomrule

\end{tabular}
\vspace{0.2cm}
\caption{Examples of opposing stances in the offensive dataset \citep{sap-etal-2020-social}. Three-ordinal ratings are used for labels, with scores from 0 to 2, with 0 as \textit{not offensive} and 2 as \textit{offensive}.}
\vspace{0.12cm}
\label{tab:offensive-annotations}
\end{wraptable}

Beyond variance, it is also crucial to assess distribution structures \citep{akhtar2021whose, van2001measuring} and whether genuine opposing opinions exist for an item. Human annotations on offensive classification in \cref{tab:offensive-annotations} suggest that disagreement can arise from different interpretations of what constitutes offensive content. In Example 1, disagreement emerges in a case that involves political satire. Some annotators interpret mocking a political figure as offensive, possibly due to perceived contempt or disrespect, whereas others regard it as legitimate political expression or humor rather than offensive content per se. In some cases, annotators label a comment as offensive due to the presence of offensive terms, even when the overall intent of the comment is not to insult. For example, in Example 3 of \cref{tab:offensive-annotations}, a comment condemning derogatory terms toward women may still be marked offensive by some because it quotes them, while others focus on the critical intent and label it non-offensive. These examples show that annotation variation is often driven by differences in how annotators weigh lexical content, speaker intent, and contextual meaning about respect and harm.

\begin{wrapfigure}{l}{0.5\textwidth}
\vspace{-0.8cm}
    \centering \includegraphics[width=\linewidth]{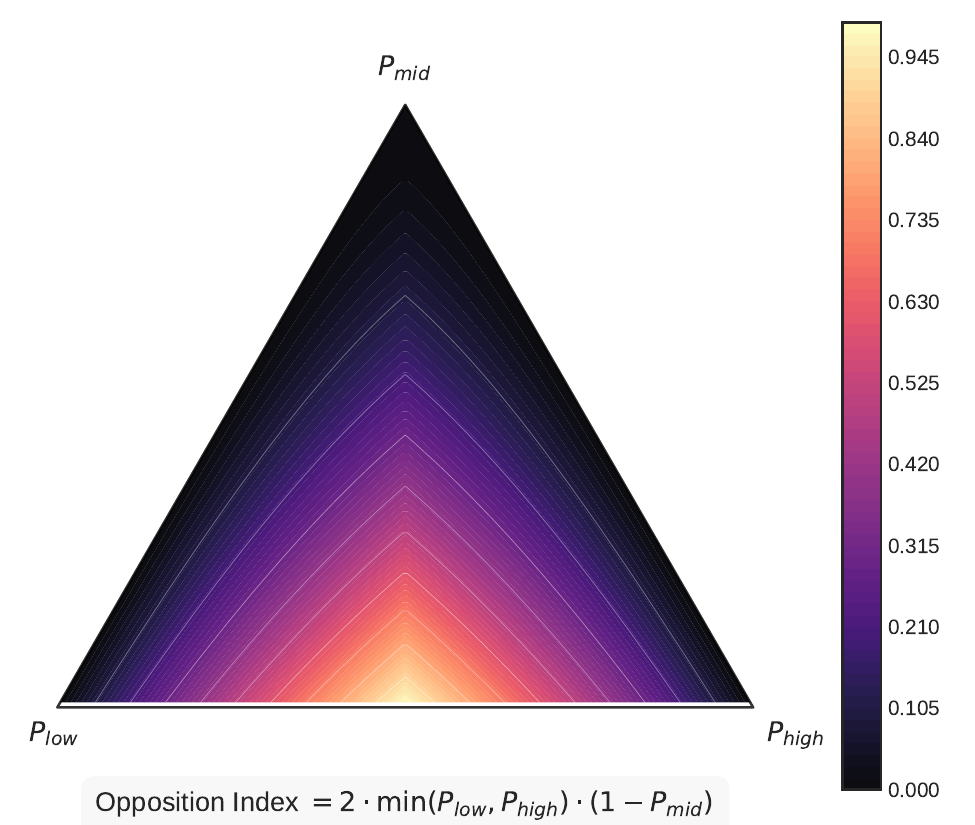}
    \vspace{0.15cm}
    \caption{Opposition Index Illustration}
\label{fig:oppo_index_illustration}
\end{wrapfigure}

\vspace{0.2cm}

Reflected in the Likert distribution, this manifests as bimodal patterns rather than a single Gaussian mode. {Traditional bimodality measures, such as
the Bimodality Coefficient (BC), or mixture-model-based modality tests, 
are typically designed for continuous distributions with sufficiently 
large sample sizes. When the number of annotators per item is small (e.g., around five), these statistics become unstable and lack statistical power, making them unsuitable for reliably detecting bimodality in item-level annotation distributions.}

\vspace{0.2cm}

To address the problem with sparse annotations per item, we propose a metric to quantify opposing stances, referred to as \textbf{opposition index}. Let the predicted or observed distribution for an item be divided into three segments: the low, mid, and high ratings of the Likert scale, with probabilities denoted as \(P_{\text{low}}, P_{\text{mid}}, \text{and }  P_{\text{high}}\). 
\noindent \(P_{\text{low}}\) corresponds to the proportion of annotators giving the lowest ratings (e.g., not toxic). \(P_{\text{high}}\) corresponds to the proportion giving the highest ratings (e.g., extremely toxic); and \(P_{\text{mid}}\) corresponds to the proportion in the middle of the scale. \\

We define the opposition stance index as:\\
\begin{equation}
\text{Index}_i = 2 \cdot \min(P_{\text{low}}, P_{\text{high}}) \cdot (1 - P_{\text{mid}})
\end{equation}

\noindent The final index value ranges from 0 to 1, with 1 indicating maximal polarization (half of the annotators select the low end and half the high end, with no intermediate ratings), with a value of 0 indicating consensus, reflected in a unimodal distribution centered in the middle or skewed toward either end of the scale, with no annotations spanning both extremes. \cref{fig:oppo_index_illustration} illustrates how the index behaves: when \(P_{\text{low}} = 0.5\) and \(P_{\text{high}} = 0.5\) with negligible \(P_{\text{mid}}\), the index reaches its maximum, reflecting clear opposing stances among annotators.

\section{Objective Functions for Likert Distribution Prediction \hspace{35pt}}\label{loss functions}

To model both annotation variation and opposing opinions discussed in the previous section, we infer the full Likert rating distribution for each item, apart from predicting a single summary statistic of variance. In this section, we propose objectives specifically designed for Likert-scale ratings, leveraging their ordinal structure rather than treating them as categorical labels.
For an item \(i\) with \(K\) Likert categories, the target distribution
is represented as:
\begin{equation}
\mathbf{p}_i = [p_{i1}, p_{i2}, \dots, p_{iK}], \text{ where } 1 < 2 < \cdots < K.
\end{equation}

\noindent where \(p_{ik}\) denotes the proportion of annotators assigning rating
\(k\) to item \(i\), and
\(\sum_{k=1}^{K} p_{ik} = 1\).
For example, in a 5-point Likert setting (ratings range from 0 to 4), if the annotations for an item are \(\{1, 1, 0, 1, 2\}\) collected from 5 annotators, the empirical distribution is represented as a vector: \(\mathbf{p}_i = [0.2, 0.6, 0.2, 0.0, 0.0].\)

\vspace{8pt}

\noindent To train the model to predict distributions, we experiment with loss functions listed below:

\paragraph{Earth Mover's Distance} (EMD), also known as the Wasserstein distance \citep{rubner2000earth}, explicitly accounts for the ordinal distance between rating categories.
EMD penalizes prediction errors proportionally to the distance between categories.

\paragraph{EMD with Mean Regularization}

We further propose a multi-task objective that combines
Earth Mover's Distance with an explicit constraint on the predicted mean rating with mean squared error. {While EMD captures
overall distributional shifts and respects the ordinal
structure of the Likert scale, it does not directly constrain
the expected rating (i.e., the mean of the distribution).
Two distributions with similar cumulative shapes
may still differ in central tendency.
To address this, we introduce an additional mean-squared error
term on the expected rating.} 






\paragraph{Ordinal Cumulative Cross Entropy}

To capture the ordinal structure of Likert-scale labels, we customize cross-entropy loss to measure the distributional difference of the Likert ratings. In our approach, a $K$-level Likert-scale problem is transformed into $K-1$ binary decisions with positive class as $y > k$. The total loss is then computed as the sum of the losses over all $K-1$ thresholds.


\paragraph{Kullback–Leibler Divergence} 
We also consider the Kullback--Leibler (KL) divergence \citep{kullback1951information} as a comparison to cumulative probability approaches and use it to quantify the dispersion from the predicted distribution to the target distribution (soft-label representation, which we refer to as KL-soft in this work).

\section{Experiments and Implementation}

This section introduces the datasets used for the experiments in this study and experiment implementation details.

\subsection{Datasets}



\renewcommand{\arraystretch}{1} 
\begin{figure}[!t]
    \centering
    \includegraphics[width=1\linewidth]{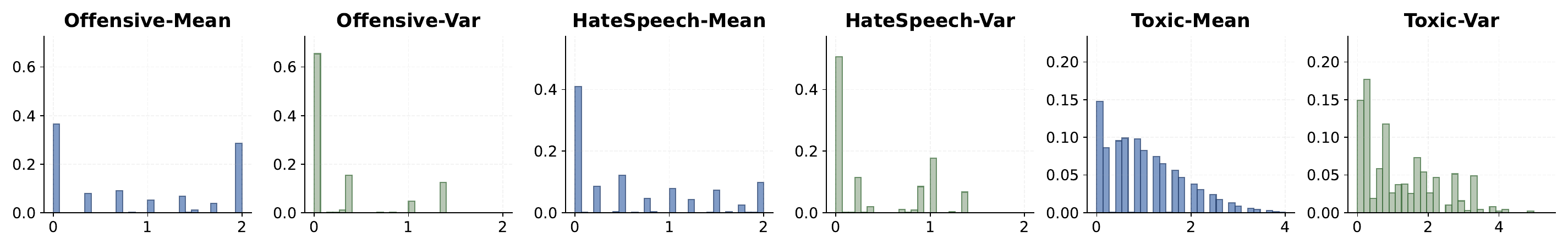}
    \vspace{0.0cm}
 \caption{Summary of Dataset Statistics: Mean and Variance. The y-axis values indicate normalized density, which accounts for the relative proportion of data points in each bin.} \label{fig:dataset_stats}\vspace{0.2cm}
\end{figure}

We conduct experiments on three subjective datasets annotated by multiple annotators. \cref{fig:dataset_stats} shows the summary of the statistics (annotation mean and variance) of three datasets.

\paragraph{Offensive Language.}
The Offensive Language dataset \citep{sap-etal-2020-social} is annotated with three categories of offensiveness: \textit{no}, \textit{maybe}, and \textit{yes}.
We map these categories to a three-point Likert scale (0 - 2). The dataset contains approximately 150k annotated items.
To ensure reliable empirical annotation distributions, we filter out items annotated by fewer than three annotators.
After filtering, the dataset comprises approximately 128.6k annotations over 35.8k unique items, with an average of {4 annotations per item}.

\paragraph{Hate Speech.} 
The hate speech dataset by \citet{kennedy2020constructing} provides graded annotations of hatefulness. Labels are provided on a three-level Likert scale, where 0 denotes non-hateful content and higher values indicate increasing hate speech severity.
Items annotated by fewer than four annotators are removed, resulting in approximately 67k annotations over 5,990 unique items, and each item is roughly annotated by {11 annotators} on average.

\paragraph{Toxicity.} Annotations in the dataset \citep{kumar2021designing} follow a five-point Likert scale ranging from \textit{not toxic} (0) to \textit{extremely toxic} (4). 
The dataset contains approximately 107.6k text instances, most of which are annotated by {5 annotators}.
We retain items with at least five annotations and merge repetitive comments, resulting in approximately 106k items.

\subsection{Implementation}
\vspace{10pt}

\paragraph{Data Splits.} Each dataset is randomly divided into training, validation, and test sets, following a 50\%, 25\%, 25\% ratio. It is partitioned at the level of distinct text instances to prevent any items from appearing in multiple data splits. 

\paragraph{Model Setting.} Models are implemented in PyTorch, and text inputs are encoded using the pretrained Sentence-Transformer model \footnote{\texttt{all-MiniLM-L6-v2}: \url{https://huggingface.co/sentence-transformers/all-MiniLM-L6-v2}}
 \citep{reimers-2019-sentence-bert}. To allow fair comparison across different prediction objectives, we keep the model architecture fixed, including input features, number of hidden layers, and layer dimensions, for all experiments. Models are trained with early stopping. Training is terminated if the validation performance does not improve for five consecutive epochs. The best-performing model on the validation set in the training history is selected for evaluation and reporting. We use the {aggregated binary distribution} (where responses greater than $(K-1)/2$ for K-class Likert are treated as the positive class) as a {baseline} for each dataset, training with binary cross entropy loss, and compare its prediction with direct variance regression and full Likert distribution prediction.

\paragraph{Evaluation Protocol.} For reliability, each experiment is repeated with five independent random splitting seeds, and the mean of the evaluation metrics is reported as the model performance.
For the computation of the opposition index, we treat rating 0 as the low value and 2 as the high value for three-class Likert ratings, and treat ratings 0 and 1 as the low-value group and ratings 3 and 4 as the high-value group for five-class Likert ratings, representing two opposing stance camps.

\begin{figure}[t]
\vspace{0.1cm}

    \centering
    \includegraphics[width=\linewidth]{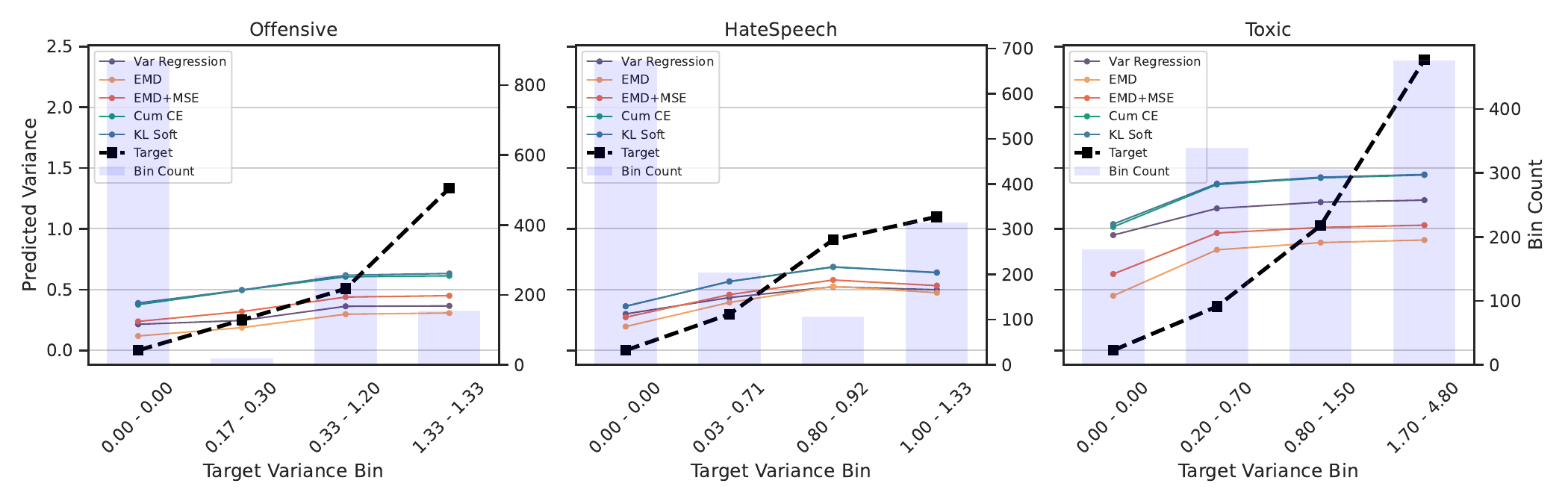}
\caption{Item-level variance grouped by target variance bins: true versus predicted variance. }
    \label{fig:variance_fig} \vspace{0.1cm}
\end{figure}

\vspace{0.3cm}

\section{Results and Discussion}

This section presents the experimental results for both annotation variance estimation (\cref{variance_result}) and opposing stance prediction (\cref{opposing_result}).

\subsection{Rating Variance Estimation}\label{variance_result}

\begin{wraptable}{l}{0.5\textwidth}{}
\centering
\small
\resizebox{\linewidth}{!}{%
\renewcommand{\arraystretch}{1.16} 
\begin{tabular}{l c c c}
\toprule
 \textbf{Model} & \textbf{Var\_MSE $\downarrow$} & \textbf{Var\_Corr $\uparrow$} & \textbf{Disagree\_F1 $\uparrow$} \\
\midrule
\rowcolor{lightblue} \multicolumn{4}{l}  {\textbf{Offensive}} \\ \midrule
Binary CE        &  0.218 {\scriptsize$\pm$ 0.003} & 0.344 & 0.5917 {\scriptsize$\pm$ 0.007} \\ \hdashline
Var Reg       &  \textbf{0.195} {\scriptsize$\pm$ 0.008} & 0.389 & \textbf{0.6112} {\scriptsize$\pm$ 0.018} \\
EMD           &  0.227 {\scriptsize$\pm$ 0.008} &\textbf{ 0.393 }& 0.5957 {\scriptsize$\pm$ 0.018} \\
EMD+MSE       &  0.206 {\scriptsize$\pm$ 0.003} & 0.386 & 0.6071 {\scriptsize$\pm$ 0.018} \\

Cum CE        &  0.248 {\scriptsize$\pm$ 0.006} & 0.371 & 0.6076 {\scriptsize$\pm$ 0.014} \\
KL Soft       &  0.251 {\scriptsize$\pm$ 0.014} & 0.372 & 0.6096 {\scriptsize$\pm$ 0.022} \\\midrule

\rowcolor{lightblue} \multicolumn{4}{l}{\textbf{Hate Speech}} \\\midrule
Binary CE        &  0.275 {\scriptsize$\pm$ 0.014} & 0.435 & 0.7275 {\scriptsize$\pm$ 0.014} \\\hdashline
Var Reg       &\textbf{  0.197} {\scriptsize$\pm$ 0.003} & 0.424 & 0.7233 {\scriptsize$\pm$ 0.016} \\
EMD           &  0.216 {\scriptsize$\pm$ 0.019} & \textbf{0.454 }& 0.7365 {\scriptsize$\pm$ 0.013} \\
EMD+MSE       &  \textbf{0.197} {\scriptsize$\pm$ 0.011} & 0.445 & 0.7313 {\scriptsize$\pm$ 0.017} \\

Cum CE        &  0.204 {\scriptsize$\pm$ 0.008} & 0.449 & \textbf{0.7408} {\scriptsize$\pm$ 0.016} \\
KL Soft       &  0.206 {\scriptsize$\pm$ 0.008} & 0.445 & 0.7371 {\scriptsize$\pm$ 0.011} \\\midrule

\rowcolor{lightblue} \multicolumn{4}{l}{\textbf{Toxic}} \\ \midrule
Binary CE        &  2.203 {\scriptsize$\pm$ 0.021} & 0.290 & 0.9185 {\scriptsize$\pm$ 0.001} \\\hdashline
Var Reg       &  \textbf{1.005} {\scriptsize$\pm$ 0.027} &\textbf{ 0.308} & 0.9185 {\scriptsize$\pm$ 0.007} \\
EMD           &  1.179 {\scriptsize$\pm$ 0.060} & 0.307 & 0.9186 {\scriptsize$\pm$ 0.008} \\
EMD+MSE       &  1.087 {\scriptsize$\pm$ 0.049} & 0.303 & 0.9187 {\scriptsize$\pm$ 0.007} \\
Cum CE        &  1.056 {\scriptsize$\pm$ 0.028} & 0.306 & 0.9191 {\scriptsize$\pm$ 0.007} \\
KL Soft       &  1.061 {\scriptsize$\pm$ 0.012} & 0.298 & 0.9191 {\scriptsize$\pm$ 0.007} \\

\bottomrule
\end{tabular}
}
\caption{Comparison of models for estimating item-level annotation variance (mean $\pm$ std).}\vspace{-14.5pt}
\label{var_pred}
\end{wraptable}

\paragraph{Variance Prediction} 
Models predict annotation variance values with reasonable accuracy. They achieve a variance MSE of around 0.2 on the 3-point Likert annotation tasks (Offensive and Hate Speech), and a variance MSE of approximately 1 for the 5-point Likert task (Toxic). 
Among all methods, directly predicting the unbiased variance using a regression model achieves the best performance. Among models that predict the Likert distribution and then compute variance, those trained with the EMD with mean regularization (EMD+MSE) achieve the second-best performance. They consistently outperform the EMD and KL-soft models.

\paragraph{Prediction across Annotation Variance Bins\vspace{0pt}} 
Apart from overall performance, we also analyze variance prediction across
bins divided based on human-annotated variance levels (see
\cref{fig:variance_fig}). 

The bin-grouped analysis reveals a similar pattern across models. All variance predictions exhibit a monotonic trend: items with near-perfect agreement are assigned the lowest predicted variance scores, and items with higher empirical variance tend to receive higher predicted variance. 

\clearpage

However, the predicted variance values tend to concentrate around a middle range.
The difference between low, medium, and high variance bins is
attenuated. 
For instance, the magnitude differences are underestimated for the highest
variance category. Models do not distinguish well between items with moderate and high variance. It may be due to the relatively smaller number of examples in these bins for the offensive and hate speech datasets.
For the lowest variance bins, predicted variance values rarely reach exactly 0, particularly for distribution-based models. As a result, the lowest bin is not as low as the target variance.

\paragraph{Spearman Correlation}
Models show a moderate positive correlation with human annotation variance, ranging from approximately 0.3 to 0.45 across three datasets. Some models (e.g., EMD) trained to predict Likert distributions often achieve higher Spearman correlations with human annotation variance compared to regression models that directly predict the variance. 
This suggests that even when variance is not explicitly
supervised, distribution-based training objectives can recover the
annotation variation structure across items. By contrast, directly regressing the variance can force the model to overfit the noise in the dataset. In contrast, predicting the full rating distribution allows the model to capture more structured patterns in annotator judgments. Rather than fitting a single summary statistic, the model learns the overall shape of the response distribution, providing a richer and more robust representation of opinion divergence. 
However, as expected, the loss function (KL-soft) that does not account for rating distances performs the worst, particularly for tasks with more Likert classes, such as the 5-rating Toxic dataset.

\paragraph{F1 Score} $F_1$ score in this study measures whether an item’s annotations exhibit rating differences (that is, whether the variance is greater than 0). Across all three datasets, the F1 scores from different models are generally very similar. 
On the Offensive dataset, the Variance Regression model achieves the highest F1 score (0.611 ± 0.018), slightly outperforming the other models. For hate speech, distribution prediction with cumulative cross-entropy loss achieves the best performance. On the Toxic dataset, all models achieve high (\textasciitilde0.92) and nearly identical F1 scores, likely because the dataset is annotated using a 5-point Likert scale, and most items have variance greater than 0. The imbalance tendency toward the class of presence of annotation variation (around 85\% items) may lead to most results at class 1 (the presence of annotation divergence). 
Overall, the models achieve strong performance in detecting variation, with $F_1$ scores exceeding 0.6 for nearly all models across the three datasets.


\subsection{Opposing Stance Prediction}\label{opposing_result}
\begin{figure*}[t!]
    \centering
\includegraphics[width=1\linewidth]{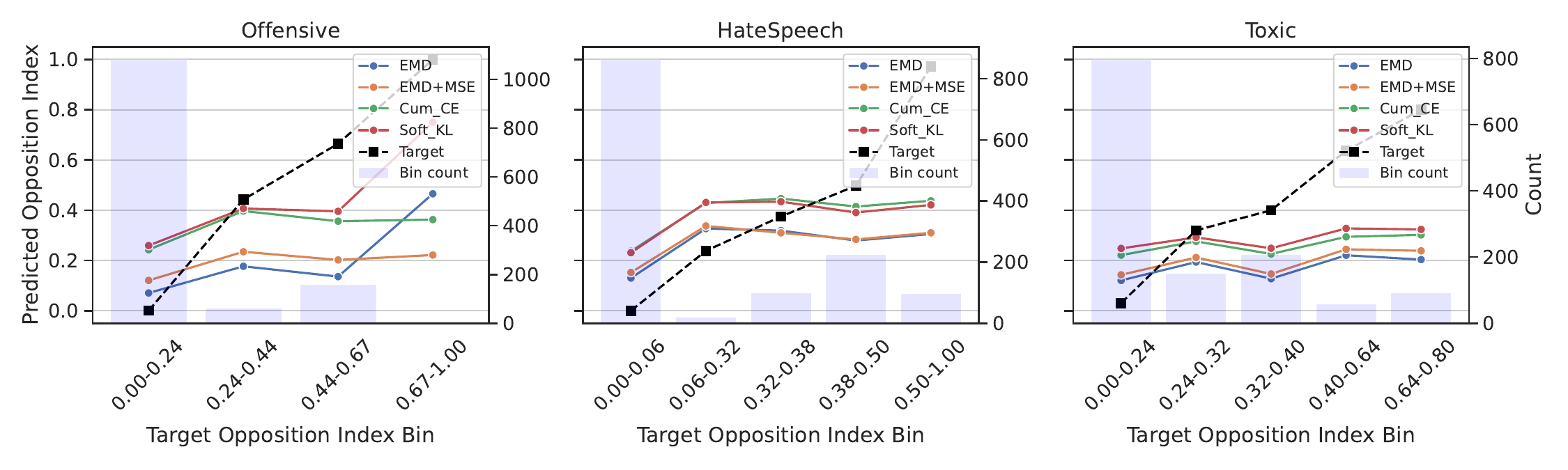}
\caption{Opposition index values binned according to target scores.}
\vspace{0.2cm}
    \label{fig:opposition_fig}
\end{figure*}
For opposing stances measured by the polarization index, we group items into intervals based on their target opposition index values and analyze the results within each interval.
Ideally, items with higher true opposition
values should also receive higher predicted scores from models.
However, this pattern does not fully hold. While models successfully distinguish
between items with no opposition (index close to 0) and items with
moderate opposition, they struggle to capture extreme
polarization. For items whose true opposition index approaches 1,
predicted values tend to remain in a mid-range (approximately 0.4),
indicating underestimation of highly polarized cases. 

\vspace{0.2cm}

Several factors may explain this phenomenon. First, as shown in
\cref{fig:opposition_fig}, the number of items in the highest opposition index bin is relatively small, which limits the model’s ability to learn stable patterns for extreme polarization when such cases are rare in the training set.
Secondly, items with a high polarization index are more prone to noise when a few annotators deviate significantly from the majority. When annotation noise inflates the opposition index, the resulting patterns may not reflect stable item features, causing the model to regress toward the mean. 
Finally, extreme opposition may partly result from other factors beyond the text itself, such as annotators' ideological differences, personal experience, or differing interpretations of the guidelines, which cannot be inferred from textual features alone. Across the three datasets, items are labeled by annotators with diverse socio-demographic backgrounds, which are not evenly distributed across items. As a result, certain influences cannot be fully inferred from item-level features alone.

\section{Conclusion}
This study investigates the extent to which annotation variation can be inferred from item-level features alone. We examine two aspects of annotation variation: estimating annotator rating variance and predicting instances that may elicit opposing human perspectives. Our results indicate a moderate positive correlation between estimated and observed annotation variance. We further show that variance estimates derived from predicted annotation distributions achieve performance comparable to variance prediction from the regression model when trained with appropriate objectives, such as Earth Mover’s Distance and its mean-regularized variant.
Beyond predicting a single summary statistic such as variance, distribution-based approaches have the potential to capture disagreement structure, such as stance opposition among annotators. To quantify this phenomenon, we propose the \textit{Opposition Index} and demonstrate its applicability across three datasets. We find that items with high opposition index values are more difficult to predict and are frequently underestimated by models.

\vspace{0.12cm}
These findings suggest practical implications for future annotation design and resource allocation. In practice, annotation effort may be allocated more efficiently by assigning more annotators to instances likely to elicit substantial opinion divergence, while reducing annotation effort on instances with strong consensus.


\section*{Limitations}

Although this paper examines the predictability of annotation variation from textual features, it cannot be assumed that the state-of-the-art encoder model, which converts texts into embeddings, perfectly captures all textual information. Prior work has identified limitations of embedding models in semantic representation \citep{lucy-gauthier-2017-distributional,zhang-etal-2024-unveiling,zhang-coltekin-2024-tubingen}. There can be information loss during the embedding process, and some linguistic cues may not be fully represented. Secondly, the number of annotations per item is limited, which limits the reliability of opinion divergence and distributional estimates. Additionally, the observed rating distributions may be sensitive to sampling noise. The datasets used in this work are crowd-sourced. Although crowd-sourced data increases annotator diversity, it also introduces additional noise, making human opinion modeling more challenging.
Finally, Likert scales are restricted to three or five categories in the datasets we experiment with. With few annotators and coarse-grained scales, the space of possible variance or distributional values becomes highly discrete. For example, when only three annotators are available, certain distribution proportions (e.g., 0.33 or 0.66) occur frequently due to combinatorial constraints rather than meaningful underlying differences. This discreteness reduces the granularity of human opinion divergence and can affect the interpretability of predicted distributions.


\clearpage
\bibliographystyle{colm2024_conference}
\bibliography{colm2024_conference}


\end{document}